\pdfoutput=1
\documentclass[11pt]{article}
\usepackage{acl}
\usepackage{times,latexsym}
\usepackage[T1]{fontenc}
\usepackage[utf8]{inputenc}
\usepackage{microtype}
\usepackage{booktabs}
\usepackage{array}
\usepackage{graphicx}
\usepackage{amsmath,amssymb}

\title{Fan Out or Relay? Ownership-Aware Routing for\\LLM Agents Querying Siloed Tables}
\author{Zhexi Feng \\
Department of Electrical and Computer Engineering \\
University of California San Diego \\
La Jolla, CA 92093, USA \\
\texttt{zhf023@ucsd.edu}}

\begin{document}
\maketitle

\begin{abstract}
Inside organizations, a question's data often sit with several owners, who query their own tables but never hand them over. An LLM coordinator can fan out, sending the question to every owner at once, or relay, querying owners in turn and reusing earlier rows. We show that neither is safe in advance, because data ownership decides which one is right. Across thirteen settings on PeopleJoin and BIRD with three model families, fan-out is more accurate in twelve on questions one owner can answer, yet answers at most 4.3\% of those joining two owners' tables. The flip follows ownership, not query shape: on BIRD two-table joins, fan-out answers 32--44\% when one owner holds both and 1--4\% otherwise. Relay wins by conditioning on returned rows, and hiding them from an otherwise identical planner cuts two-owner accuracy from 23.2\% to 5.1\%. Since ownership is public, we propose Ownership-Aware Routing (OAR), which drafts a schema-only SQL query, counts the owners whose tables it joins, and routes the question before any private row is read. In ten settings OAR is never significantly less accurate than the best fixed protocol, and beats every fixed protocol on PeopleJoin, by up to 9.3 points on held-out organizations.
\end{abstract}

\section{Introduction}

\begin{figure*}[t]
\centering
\includegraphics[width=\textwidth]{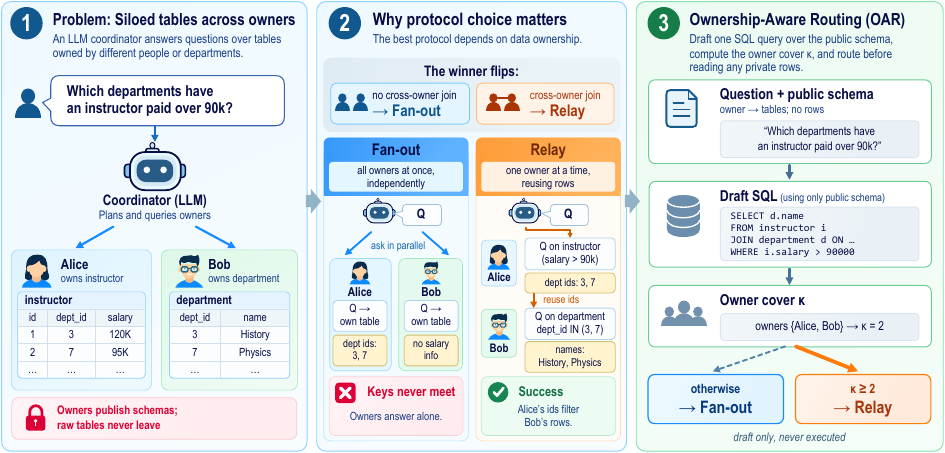}
\caption{(1)~A coordinator answers a question over tables held by different owners, who publish their schemas but keep their rows. (2)~When the question joins tables of two owners, fan-out fails because Alice's department ids never reach Bob, while relay places them in its query to Bob. (3)~OAR drafts one SQL query over the public schema, counts the owners whose tables it joins (its owner cover $\kappa$, Equation~\ref{eq:cover}), and routes the question before any private row is read.}
\label{fig:overview}
\end{figure*}

In a university, a hospital or a company, many questions put to an LLM assistant cannot be answered from any single place. Course records may sit with the registrar and salary data with human resources, and each owner is willing to run a query over its own tables but not to hand the tables over. PeopleJoin \citep{jhamtani-etal-2025-llm} and iAgents \citep{liu2024iagents} study agents that gather facts held by other users, and tool interfaces now expose such data one query at a time \citep{schick2023toolformer,anthropic2024mcp}. A coordinator must decide whom to ask, what to ask and in what order, while every returned row competes for a limited communication budget. How it asks can decide whether a question is answered at all.

Two ways of asking recur in existing systems. A coordinator can \emph{fan out}, sending the question to all relevant owners in parallel and merging their answers, much as parallel function calling runs independent tool calls at once \citep{kim2024llmcompiler}. It can also \emph{relay}, querying one owner at a time so that each query can use rows returned earlier (Figure~\ref{fig:overview}). Relaying mirrors interleaved reasoning and retrieval \citep{yao2023react,trivedi2023ircot} and the semi-join strategies of distributed databases, which ship join keys instead of whole relations \citep{bernstein1981semijoins,kossmann2000state}.

Prior work explained the success of such collaboration from two angles. Multi-agent studies optimized the structure of communication, assigning roles, wiring agents into graphs and pruning redundant messages \citep{hong2024metagpt,zhuge2024gptswarm,qian2025scaling,zhang2025cut}. Adaptive methods matched the strategy to the question instead, choosing between one-step and iterative retrieval by question complexity \citep{jeong2024adaptiverag} or learning a collaboration mode per query \citep{yue-etal-2025-masrouter}. Both lines assumed that all agents share one view of the data. On PeopleJoin itself, an agent that sends the user's question once to every member trailed a reactive agent in aggregate \citep{jhamtani-etal-2025-llm}. None of these accounts explains why the same protocol succeeds on one question and fails on the next over the same organization's data.

We find that the missing variable is data ownership, namely whether answering a question requires joining tables held by different owners. On PeopleJoin, and on BIRD \citep{li2023bird} with its tables distributed over owners, fan-out is typically the more accurate protocol on questions that a single owner can answer. Once the needed tables span two owners, the ranking flips, and fan-out answers at most 4.3\% of the questions while relay answers up to 47.9\% (Figure~\ref{fig:crossover}). In all thirteen settings we evaluate, relay's advantage over fan-out is 16 to 57 points larger on two-owner than on single-owner questions, across three model families and on 600 questions never used during development. The flip follows ownership rather than query shape. BIRD deals tables to owners by hashing their names, and on joins of two tables fan-out answers 32--44\% when one owner holds both and 1--4\% when two owners do. When owners merely hold different rows of one table, no join is needed and fan-out usually keeps its lead.

The cause is mechanical. A join across owners needs the keys held by one owner to select the matching rows of the other. Owners that answer independently never see those keys, and a relay planner that cannot see the rows already returned loses most of its advantage on both benchmarks.

Ownership is decided by who holds which table, and that information is public even when the rows are not. This lets the coordinator predict the better protocol before it touches any private data, and we turn the observation into Ownership-Aware Routing (OAR). OAR asks the model for one draft SQL query over the public schema, maps the tables in the draft to their owners, and sends the question to relay when the draft needs two or more owners and to fan-out otherwise. The draft is never executed. It only exposes the question's ownership structure, at the cost of one model call and without any training.

Across ten settings, OAR is never significantly less accurate than the best fixed protocol chosen in hindsight. On PeopleJoin it exceeds that protocol for all three model families, by 2.4--3.3 points, and by 7.3--9.3 points on 108 organizations that played no part in its design. On BIRD, where most questions span several owners, OAR improves fan-out by 14--19 points and stays within 2.7 points of the strongest fixed protocol. When the measured accuracies are reweighted to any mix of single- and two-owner questions, OAR never falls more than 3.8 points below the better of fan-out and relay, while fan-out can fall 46.5 points behind.

Our contributions are threefold:
\begin{itemize}
\item We identify joins across owners as the variable that decides whether fan-out or relay answers a question, in thirteen settings over 134 databases and with the query shape held fixed.
\item We trace the reversal to conditioning on returned rows: hiding them from a relay planner erases most of its advantage, and planning all queries in advance recovers only part of it.
\item We propose OAR, a one-call router that reads ownership from a schema-only draft and outperforms every fixed protocol on PeopleJoin as well as routing on question difficulty or on the model's own choice of protocol.
\end{itemize}
Sections~\ref{sec:setup} to~\ref{sec:experiments} define the task, the protocols and the setup, Section~\ref{sec:results} presents the findings and Section~\ref{sec:oar} the router.

\section{Problem Setup}\label{sec:setup}

An organization consists of owners $o_1,\dots,o_m$, and owner $o_i$ holds a set of tables $\mathcal{T}_i$ (Figure~\ref{fig:overview}, panel~1). A coordinator receives a natural-language question together with a public catalog that lists, for every owner it can reach, the names and columns of the tables it holds, but none of their rows. The coordinator may send an owner the question or a read-only SQL query, and the owner runs a query on its own tables and returns the result rows. We assume that an owner may see the literal values, such as join keys, that the coordinator copies into a query to it (see Ethical Considerations). The rows that owners return, together with the queries and provenance that travel with them, are charged against a per-question budget of $B$ tokens, with $B=2048$ unless stated otherwise. The coordinator finally returns a set of rows or declares the question unanswerable. An answer is correct when its rows equal the reference rows as a multiset, and an unanswerable question counts as correct only when the coordinator abstains.

To describe how the evidence for a question is distributed, we use the \emph{owner cover} of a SQL query. For a query $Q$ that references the set of tables $T(Q)$, it is the smallest number of owners whose tables jointly contain $T(Q)$:
\begin{equation}
\kappa(Q)=\min_{S\subseteq\{1,\dots,m\}}\bigl\{|S| : T(Q)\subseteq\textstyle\bigcup_{i\in S}\mathcal{T}_i\bigr\}.
\label{eq:cover}
\end{equation}
Evaluated on a question's reference SQL, $\kappa$ separates questions whose tables one owner holds ($\kappa=1$) from \emph{two-owner} questions ($\kappa=2$) and from questions that need three or more owners. A table may also be split by rows across owners, as PeopleJoin does for one table in many of its organizations. Combining such rows takes a union, to which each holder contributes its rows independently, so in Equation~\ref{eq:cover} any holder counts as holding the table and $\kappa$ counts the owners whose tables a query must join. Among questions with $\kappa=1$, we call those that read a split table \emph{row-split} and the others \emph{single-owner}, since one owner then holds every needed row. No protocol sees the reference query, so the reference cover serves only for analysis. OAR (Section~\ref{sec:oar}) applies the same definition to a draft that the coordinator writes itself.

\section{Fan-out and Relay}\label{sec:method}

The two protocols start from the same contacted owners and the same budget, and they differ in how they query those owners (Figure~\ref{fig:overview}, panel~2). On PeopleJoin the coordinator contacts the asking user and the three colleagues whose public descriptions best match the question under BM25. Relay may also follow up to two referral messages found in these owners' documents. On BIRD every owner is contacted. Appendix~\ref{app:protocols} gives the prompts and limits.

\paragraph{Fan-out.} Each contacted owner receives the question and its own schema, writes one SQL query that computes the most useful local partial result, and returns up to 100 rows (8{,}000 bytes). The results of the other owners are packed into the budget and passed, together with the asking user's own result, to a receiver call that composes the answer. No owner sees another owner's rows, which lets all owners work in parallel.

\paragraph{Relay.} A planner queries owners one at a time. At every step it sees the question, the owners' schemas, and the queries and rows obtained so far, and it either sends one owner the next SQL query or stops. It is instructed to obtain a small filtered set of join keys first and to place those literal keys in its query to the next owner. The planner issues at most five queries, each returning up to 20 rows (1{,}600 bytes). A receiver call composes the answer from executed, untruncated results only, and it sees the SQL that produced each result. Relay attaches this provenance at the end and reserves a fixed 256-token share of the budget for it. \emph{Relay+} instead attaches the SQL to each result as it is returned and charges it immediately, so no share has to be reserved.

\section{Experimental Setup}\label{sec:experiments}

\paragraph{Evaluation pools.} Table~\ref{tab:pools} summarizes the six pools. The two benchmarks complement each other. PeopleJoin-QA \citep{jhamtani-etal-2025-llm} distributes the tables of Spider databases \citep{yu2018spider} over the members of simulated organizations, while on BIRD we assign tables to owners by a rule that ignores the questions. We give every PeopleJoin member an owner-local SQL interface to the tables in their documents. PJ-8a and PJ-8b are disjoint question samples from the same eight organizations, on which the routing rule was developed, and PJ-7 contains 511 questions from seven further organizations. PJ-Unseen holds 300 questions from 108 organizations that were never used during development. For BIRD \citep{li2023bird} we distribute the tables of each database over up to four owners with a fixed rule that hashes the table names and use the 500 Mini-Dev questions. BIRD-Unseen adds 300 development-set questions outside Mini-Dev (Appendix~\ref{app:pools}). Every pool was drawn from public identifiers before any model saw it, and the routing rule of Section~\ref{sec:oar} was fixed before the PJ-7, BIRD and unseen evaluations on which routers are compared.

\begin{table}[t]
\centering\small
\setlength{\tabcolsep}{1.9pt}
\begin{tabular*}{\columnwidth}{@{\extracolsep{\fill}}lrrrrrl}
\toprule
Pool & DBs & Qs & 1-own. & Split & 2-own. & Models \\
\midrule
PJ-8a & 8 & 96 & 41 & 2 & 36 & DS \\
PJ-8b & 8 & 494 & 275 & 14 & 138 & DS, So \\
PJ-7 & 7 & 511 & 203 & 45 & 134 & DS, So, Qw \\
PJ-Unseen & 108 & 300 & 147 & 29 & 71 & DS, Qw \\
BIRD & 11 & 500 & 130 & -- & 293 & DS, So, Qw \\
BIRD-Unseen & 11 & 300 & 68 & -- & 171 & DS, Qw \\
\bottomrule
\end{tabular*}
\caption{Evaluation pools. Single-owner, row-split and two-owner counts use the reference cover $\kappa$ (Section~\ref{sec:setup}). The other questions need three or more owners, are unanswerable or have references whose tables could not be resolved. DS, So, Qw: DeepSeek, Sonnet, Qwen.}
\label{tab:pools}
\end{table}

\paragraph{Models.} The three model families cover an open-weight model run locally and two API models from different providers. We run DeepSeek-V4.1-Flash \citep{deepseek2026v41flash} through its API, with two runs per question on PJ-8, PJ-7 and BIRD and one on the unseen pools, Claude Sonnet 4 \citep{anthropic2025claude4} through OpenRouter with one run, and Qwen3.8-27B \citep{qwen2026qwen38} locally with 5-bit weights and one run. All calls use temperature 0 and at most 1{,}024 output tokens. Fan-out, relay and the router OAR (Section~\ref{sec:oar}) run on every pool except the two PJ-8 samples, which compare fan-out and relay. Relay+ and OAR+ run with DeepSeek on PJ-7 and with DeepSeek and Qwen on BIRD and both unseen pools.

\paragraph{Statistics.} Scores are averaged over runs within each question. Confidence intervals are 95\% percentile intervals from 10{,}000 paired bootstrap resamples of questions. For the interaction between protocol and ownership, the single-owner and two-owner strata are resampled separately, and Appendix~\ref{app:stats} also resamples whole databases. Code, prompts and per-question results are included in the supplementary material.

\section{Ownership Decides Which Protocol Wins}\label{sec:results}

\subsection{The winner flips with ownership}\label{sec:reversal}

\begin{figure*}[t]
\centering
\includegraphics[width=\textwidth]{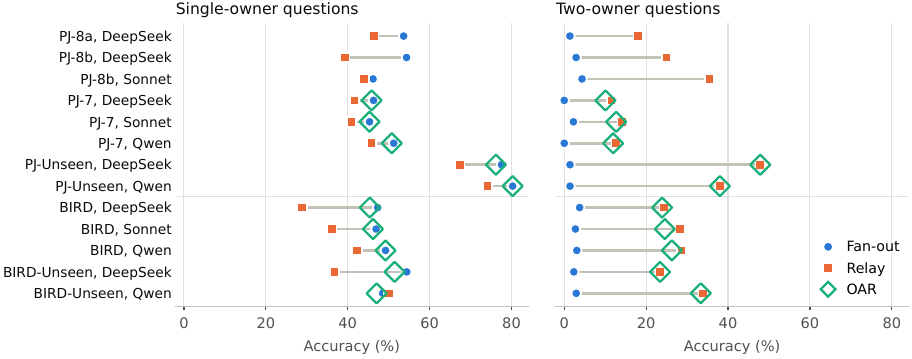}
\caption{Accuracy of fan-out and relay on single-owner questions (left) and two-owner questions (right) in all thirteen settings. Diamonds mark OAR in the ten settings where it runs. Fan-out leads on the left and collapses on the right, while OAR stays close to fan-out on the left and to relay on the right.}
\label{fig:crossover}
\end{figure*}

Fan-out wins when one owner holds the needed tables and collapses when they span two owners (Figure~\ref{fig:crossover}). On single-owner questions, fan-out is more accurate by 2.2 to 15.1 points on PeopleJoin and by 6.9 to 18.5 points in four of the five BIRD settings, while relay leads by 1.5 points on BIRD-Unseen with Qwen. On two-owner questions, fan-out answers 0.0--4.3\% of the questions correctly in every setting, while relay reaches 11.6--47.9\%.

\paragraph{The reversal holds in every setting.} We summarize the reversal by the interaction between protocol and ownership, defined as relay's advantage over fan-out on two-owner questions minus its advantage on single-owner questions. The interaction ranges from +16.2 points on PJ-7 with DeepSeek to +56.7 points on PJ-Unseen with DeepSeek, and its 95\% interval excludes zero in all thirteen settings (Appendix~\ref{app:results}). Even on BIRD-Unseen with Qwen, the one setting where relay leads on single-owner questions, it is +29.5 points. The effect is also consistent inside individual databases: the interaction is positive in 74 of the 78 per-database estimates available for PJ-8, PJ-7 and BIRD.

\begin{table}[t]
\centering\small
\setlength{\tabcolsep}{3.4pt}
\begin{tabular*}{\columnwidth}{@{\extracolsep{\fill}}lrrrrr}
\toprule
 & \multicolumn{2}{c}{One owner} & \multicolumn{2}{c}{Two owners} & \\
\cmidrule(lr){2-3}\cmidrule(lr){4-5}
Setting & FO & RL & FO & RL & Diff. \\
\midrule
BIRD, DS & 36.4 & 25.0 & 4.2 & 25.9 & +33.0 \\
BIRD, So & 31.8 & 27.3 & 3.0 & 28.9 & +30.4 \\
BIRD, Qw & 36.4 & 27.3 & 2.7 & 30.0 & +36.5 \\
BIRD-Uns., DS & 44.4 & 22.2 & 1.3 & 23.9 & +44.9 \\
BIRD-Uns., Qw & 38.9 & 38.9 & 3.1 & 34.6 & +31.4 \\
\bottomrule
\end{tabular*}
\caption{Accuracy (\%) on questions whose reference SQL joins exactly two tables, by whether the hash-based partition gives both tables to one owner (44 questions on BIRD, 18 on BIRD-Unseen) or to two (263 and 159). FO: fan-out, RL: relay. Diff.: fan-out's lead with one owner minus its lead with two, in points, with every 95\% interval above zero.}
\label{tab:natural}
\end{table}

\paragraph{The flip follows ownership, not query shape.} A two-owner question also joins at least two tables, so the flip could reflect query shape rather than ownership. BIRD separates the two: its owners are dealt tables by hashing the table names, so whether the two tables of a join fall to one owner is unrelated to the question. Among questions whose reference SQL joins exactly two tables (Table~\ref{tab:natural}), when one owner holds both tables, fan-out answers 32--44\% of them and is never less accurate than relay. When two owners hold them, fan-out answers 1--4\%, while relay answers 22--39\% in both cases. PeopleJoin offers the converse case. Many of its organizations split the rows of one table between two members, so some questions need two owners but no join. On these row-split questions fan-out leads relay in six of the eight PeopleJoin settings, by 2.2 to 44.8 points, including all five settings on PJ-7 and PJ-Unseen (Table~\ref{tab:strata-main} in Appendix~\ref{app:results}). The protocol is decided by a join across owners, not by the number of tables or owners involved.

\subsection{Relay wins by conditioning on returned rows}\label{sec:mechanism}

A two-owner question requires matching rows across owners, because the rows of one owner decide which rows of the other are relevant (Figure~\ref{fig:overview}). Fan-out owners write their queries without seeing each other's results, so neither side can restrict its rows to the matching keys and the rows the join needs rarely reach the receiver, which abstains on 78--90\% of the two-owner questions (Appendix~\ref{app:results}). Relay can put the keys it has already received into its next query.

\begin{figure}[t]
\centering
\includegraphics[width=\columnwidth]{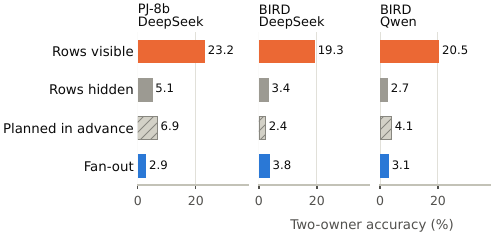}
\caption{Two-owner accuracy in the mechanism experiments. The first two bars show the Relay+ planner without its key-passing instruction, with the returned rows visible or hidden. The planned variant writes all of relay's queries before any row returns. With Sonnet on PJ-8b the comparisons point the same way with smaller gaps.}
\label{fig:mechanism}
\end{figure}

\paragraph{Hiding returned rows removes relay's advantage.} We test this account on PJ-8b with DeepSeek. We take the Relay+ planner without its key-passing instruction and run it in two conditions that differ in one switch: it either sees the rows returned so far or only learns that its queries executed. Hiding the rows lowers accuracy on two-owner questions from 23.2\% to 5.1\% (Figure~\ref{fig:mechanism}, $-18.1$ points, 95\% CI $[-24.6,-12.0]$) and leaves single-owner accuracy unchanged at 42.9\% versus 43.3\%. The planner without rows even issues 14\% more calls, so the loss does not come from fewer queries. On BIRD, hiding the rows costs the planner 17.7 points of two-owner accuracy with Qwen and 15.9 points with DeepSeek, down to the level of fan-out, while single-owner accuracy again stays unchanged (Appendix~\ref{app:mechanism}). Relay's advantage on two-owner questions comes from conditioning later queries on earlier rows.

\paragraph{Successful relay runs pass keys.} Traces from all ten evaluation settings show the same behavior (Table~\ref{tab:keys} in Appendix~\ref{app:results}). We say that a relay run passes a key when a later query to one owner contains a literal value that a different owner returned earlier, as in the example of Table~\ref{tab:example}. On two-owner questions, relay passes a key in 45--92\% of its runs. These runs produce 93--98\% of relay's correct two-owner answers on BIRD and 85--91\% on PJ-Unseen. With DeepSeek and Qwen on BIRD and PJ-Unseen, runs that pass a key are 3.1 to 8.6 times as accurate as runs that do not.

\paragraph{Planning is not enough.} A coordinator could instead plan every owner's query from the public schema and run them in parallel. Our planned variant of relay writes up to five queries, including the join-key columns that a cross-owner answer needs, before any row returns, and keeps everything else of relay (Appendix~\ref{app:protocols}). On the two-owner questions of PJ-8b it answers 6.9\% with DeepSeek (Figure~\ref{fig:mechanism}) and 28.3\% with Sonnet, 18.1 and 7.2 points below relay (Figure~\ref{fig:crossover}). On BIRD it answers 4.1\% with Qwen and 2.4\% with DeepSeek, within 1.4 points of fan-out and 24.6 and 22.1 points below relay. On single-owner questions it trails fan-out by 14 to 35 points. Planning helps only as far as the planner can anticipate the keys, whereas relay reads them from the rows that come back.

\paragraph{Limits, reach and budget do not explain the collapse.} Fan-out has the larger row limit and, on PeopleJoin, reaches fewer owners. Yet it answers at most 5.2\% of the two-owner questions whose owners are all among its initial contacts, against 10.3--55.6\% for relay. Restricted to questions whose reference answer has at most 20 rows, relay's per-query limit, the interaction still ranges from $+14.6$ to $+55.9$ points, with every interval above zero. On BIRD both protocols contact every owner, and the interaction still ranges from $+29.5$ to $+39.1$ points. Halving or quadrupling $B$ on BIRD with Qwen keeps the interaction between $+31.7$ and $+34.3$ points (Appendix~\ref{app:results}).

\section{Ownership-Aware Routing}\label{sec:oar}

\subsection{Routing by a schema-only draft}\label{sec:oar-method}

Section~\ref{sec:results} showed that fan-out fails on questions that join tables of different owners, while relay trails it on most questions that one owner can answer. Which owners a question needs is visible in the public catalog, so the better protocol can be chosen before any private row is read. Ownership-Aware Routing (OAR) makes this choice with one extra call before any owner is contacted. The model receives the question and the public catalog of the owners that relay can reach and is asked for one read-only SQL query $\hat{Q}$ whose result contains exactly the requested columns and rows, or an empty query if no safe answer exists. OAR extracts the tables named in the draft's \texttt{FROM} and \texttt{JOIN} clauses, maps them to owners through the catalog, and computes the draft's owner cover $\kappa(\hat{Q})$ with Equation~\ref{eq:cover}. It routes the question to relay if $\kappa(\hat{Q})\geq 2$. The rule has no parameter to tune, since a join across owners involves at least two of them. All other questions go to fan-out, including those whose draft is empty, malformed, or names a table that no reachable owner holds. The draft is stored before routing and is never executed or shown to the chosen protocol, so OAR reads no private rows. We write OAR+ for the same router with Relay+ as its relay branch.

\subsection{OAR is robust across settings}\label{sec:main}

\begin{table*}[t]
\centering\small
\begin{tabular*}{\textwidth}{@{\extracolsep{\fill}}lrrrrrrrr}
\toprule
 & \multicolumn{3}{c}{Fixed protocols} & \multicolumn{3}{c}{Other routers} & \multicolumn{2}{c}{Ownership routing} \\
\cmidrule(lr){2-4}\cmidrule(lr){5-7}\cmidrule(lr){8-9}
Setting & Fan-out & Relay & Relay+ & Self & Ask & Difficulty & OAR & OAR+ \\
\midrule
PJ-7, DeepSeek & 39.5 & 36.9 & 38.1 & 38.6 & 39.7 & -- & \textbf{42.0}$^\dagger$ & 41.4 \\
PJ-7, Sonnet & 39.3 & 31.1 & -- & -- & -- & -- & \textbf{42.7}$^\dagger$ & -- \\
PJ-7, Qwen & 41.3 & 41.5 & -- & 42.3 & 43.4 & -- & \textbf{44.2}$^\dagger$ & -- \\
PJ-Unseen, DeepSeek & 52.7 & 54.0 & 53.0 & 60.9 & 58.3 & -- & \textbf{63.3}$^\dagger$ & 61.3 \\
PJ-Unseen, Qwen & 55.7 & 57.7 & 56.7 & 62.7 & 60.0 & -- & \textbf{65.0}$^\dagger$ & \textbf{65.0} \\
\midrule
BIRD, DeepSeek & 14.5 & 25.2 & 30.3 & 27.4 & 28.5 & 20.8 & 28.8 & \textbf{31.6} \\
BIRD, Sonnet & 13.8 & \textbf{29.2} & -- & -- & -- & 22.2 & \textbf{29.2} & -- \\
BIRD, Qwen & 14.6 & 31.6 & \textbf{33.0} & 27.4 & 31.4 & 23.6 & 31.4 & 31.4 \\
BIRD-Unseen, DeepSeek & 14.0 & 24.3 & 30.7 & 25.7 & 27.3 & 20.0 & 28.0 & \textbf{32.0} \\
BIRD-Unseen, Qwen & 13.0 & 33.0 & \textbf{33.3} & 30.7 & 32.3 & 26.0 & 32.0 & 32.0 \\
\bottomrule
\end{tabular*}
\caption{Accuracy (\%) of the fixed protocols and the routers. Each router picks the saved fan-out or relay run of a question by the model's choice (Self), by the owners the model says the question needs (Ask), by BIRD's gold difficulty labels (Difficulty) or by the draft's owner cover (OAR). Bold marks the highest accuracy in each row, and $^\dagger$ marks OAR when its 95\% interval against the best fixed protocol lies above zero (Appendix~\ref{app:results}).}
\label{tab:main}
\end{table*}

In no setting is OAR significantly less accurate than the most accurate fixed protocol of that setting, chosen in hindsight (Table~\ref{tab:main}). On the five PeopleJoin settings it is more accurate than every fixed protocol, leading the best one by 2.4 to 3.3 points on PJ-7 and by 9.3 and 7.3 points on PJ-Unseen with DeepSeek and Qwen, and every 95\% interval of these differences lies above zero (Table~\ref{tab:contrasts} in Appendix~\ref{app:results}). Because OAR uses, for every question, the fan-out or relay run of that same question, each difference from a fixed protocol comes from routing alone.

\paragraph{OAR keeps the strength of each protocol.} PeopleJoin mixes question types (Table~\ref{tab:pools}), and a fifth of the PJ-7 questions cannot be answered from the organization's data. Figure~\ref{fig:crossover} and Table~\ref{tab:strata-main} in Appendix~\ref{app:results} show that OAR keeps fan-out's accuracy on single-owner, row-split and unanswerable questions and most of relay's accuracy on two-owner questions, a combination that no fixed protocol achieves. With DeepSeek on PJ-7, for instance, OAR abstains correctly on 92.5\% of the unanswerable questions, against 78.3\% for relay, and it answers 10.1\% of the two-owner questions, against none for fan-out.

\paragraph{On BIRD, OAR follows relay.} More than 70\% of the BIRD questions need several owners, and OAR routes most of them to relay. It improves on fan-out by 14.0 to 19.0 points in the five BIRD settings and differs from the best fixed protocol by 0.0 to $-2.7$ points, with every 95\% interval of the difference including zero. OAR+, which routes to Relay+, is the most accurate protocol with DeepSeek on both BIRD pools, at 31.6\% and 32.0\% (Appendix~\ref{app:routing}).

\paragraph{OAR is robust to the question mix.} Reweighting the stratum accuracies predicts how each protocol would fare on any mix of single-owner and two-owner questions (Figure~\ref{fig:mix} in Appendix~\ref{app:results}). Across all ten settings and all mixes, OAR is never more than 3.8 points below the better of fan-out and relay, whereas fan-out can fall 46.5 points and relay 18.5 points behind. At intermediate mixes OAR exceeds both in nine of the ten settings, by up to 8.5 points on BIRD and 7.3 points on PJ-Unseen with DeepSeek, because it keeps the better protocol on each kind of question.

\paragraph{OAR costs about as much as relay.} Including its draft call, OAR costs 0.9 to 1.2 times as much as relay and 1.4 to 1.9 times as much as fan-out in the four API settings on PJ-7 and BIRD. Relay makes 3.4 to 4.6 model calls per question in sequence, against two rounds for fan-out with parallel owners and one more call for OAR's draft. In our implementation, which queries fan-out's owners in turn, relay took about as long as fan-out with the API models and up to 1.8 times as long with Qwen (Appendix~\ref{app:routing}).

\subsection{Why routing by a draft works}\label{sec:analysis}

The ablations below reroute the same saved runs by the reference cover, by other routing signals and by drafts from a smaller model, to test whether the draft reveals ownership and how much better knowledge of ownership could add.

\paragraph{Ownership is legible in the draft.} When the draft needs two or more owners, the reference query does as well in 88--99\% of the cases, and the draft identifies 59--91\% of all multi-owner questions ($\kappa\geq 2$, Table~\ref{tab:analysis} in Appendix~\ref{app:results}). High precision keeps single-owner questions with fan-out. With drafts from Llama-3.1-8B, OAR is not significantly less accurate than the best fixed protocol in any PeopleJoin setting and stays 10 to 15 points above fan-out on BIRD (Appendix~\ref{app:routing}).

\paragraph{OAR comes close to an ownership oracle.} An oracle that routes every question by its reference cover shows what perfect knowledge of ownership would achieve. On PJ-7 this oracle improves on the best fixed protocol by 3.0--4.1 points, and on PJ-Unseen by 11.3 and 7.7 points with DeepSeek and Qwen. OAR attains 81--96\% of these gains from the draft alone, and perfect knowledge of ownership, however obtained, would add at most 2.4 points in any of the ten settings. On BIRD, even an oracle with the stronger relay executor adds at most 2.8 points over the best fixed protocol, because relay already handles the multi-owner majority.

\paragraph{Owner cover against other routing signals.} Routing the same draft on whether it references at least two tables, or on whether it contains a JOIN, stays within 1.4 points of OAR in the six settings where these variants run. The proxies nearly coincide with owner cover on PeopleJoin, where every member holds at most one table. They part on the 44 BIRD questions that join two tables of one owner, where fan-out is the better protocol (Table~\ref{tab:natural}) and OAR is 4.5 to 9.1 points more accurate than routing on table count. Asking the model which owners a question needs fails on the converse case, sending 47--90\% of the row-split questions, which need two owners but no join, to relay. Letting the model choose a protocol trails OAR by 1.3 to 4.0 points, and routing by BIRD's difficulty labels by 6.0 to 8.0 points (Table~\ref{tab:main}). Owner cover tracks the join that key passing serves, which none of these signals captures.

\paragraph{A retry after fan-out helps only on BIRD.} Two-owner questions that the draft misses are answered at most 10\% of the time under fan-out, against 15--52\% for those it sends to relay. Retrying with relay whenever fan-out abstains adds 0.7 to 2.2 points to OAR on BIRD, but changes PeopleJoin accuracy by $-8.2$ to $+1.3$ points, because fan-out's abstention there is often the correct answer to an unanswerable question (Appendix~\ref{app:routing}).

\section{Related Work}\label{sec:related}

\paragraph{Multi-agent collaboration.} LLM-based multi-agent systems assign agents roles and conversation patterns \citep{li2023camel,hong2024metagpt,wu2024autogen}, let them debate \citep{du2024debate}, and learn or prune the graphs along which they communicate \citep{zhuge2024gptswarm,qian2025scaling,zhang2025cut}. \citet{lin-etal-2026-agentask} trace failures to the messages passed between agents \citep[see also][]{guo2024survey}. These systems share the task input and study how agents should talk. We study agents that hold different data, where the flow of rows between them decides whether a question can be answered at all.

\paragraph{Collaboration under information asymmetry.} iAgents \citep{liu2024iagents} let agents exchange their users' private information to complete shared tasks. Privacy studies show that models disclose context they should withhold \citep{mireshghallah2024confaide}, which motivates keeping rows with their owners. PeopleJoin \citep{jhamtani-etal-2025-llm} evaluates agents that gather information from the members of an organization and finds that a reactive agent outperforms one that messages every member once. ECHO improves such agents by learning online from their trajectories \citep{hu2025echo}. We build on PeopleJoin's organizations and show that the ranking of these two styles, which our relay and fan-out formalize, depends on whether each question joins data across owners.

\paragraph{Adaptive execution strategies.} Sequential methods interleave reasoning with tool calls or retrieval \citep{yao2023react,press2023selfask,trivedi2023ircot} or solve subproblems in order \citep{zhou2023leasttomost}, and LLMCompiler runs independent calls in parallel \citep{kim2024llmcompiler}. FanOutQA \citep{zhu-etal-2024-fanoutqa} applies the term fan-out to questions over many documents rather than to a protocol. Adaptive-RAG selects a retrieval strategy from predicted question complexity \citep{jeong2024adaptiverag}, RouteLLM selects between models \citep{ong2025routellm}, and MasRouter learns a collaboration mode, roles and models for each query \citep{yue-etal-2025-masrouter}. OAR is a training-free strategy router whose signal, the ownership that a schema-level draft exposes before any data is accessed, outperforms BIRD's difficulty labels, and perfect knowledge of ownership, however obtained, would add at most 2.4 points (Section~\ref{sec:analysis}).

\paragraph{Text-to-SQL and distributed query processing.} Spider and BIRD \citep{yu2018spider,li2023bird} evaluate SQL generation over single databases, and multi-agent pipelines such as MAC-SQL divide it among specialized roles \citep{wang2025macsql}. CSMA splits the schema of one database across agents that keep their parts private and cooperate to write a single query \citep{wu2024csma}. In our setting the catalog is public and the rows are private, so we study how to query the owners. Distributed and federated database systems have long chosen between shipping data and semi-join reduction according to where relations reside \citep{bernstein1981semijoins,sheth1990federated,kossmann2000state}. OAR brings this placement-aware choice to LLM agents that can neither see nor move the owners' tables.

\section{Conclusion}

An LLM coordinator that queries several data owners must choose how to ask them, and data ownership decides which protocol is right. Fan-out usually leads when one owner holds the needed tables but fails once a question joins tables of two owners, because only relay conditions its queries on the rows already returned. Since who holds which table is public, OAR reads ownership from one schema-only draft and routes each question before any private row is read. In ten settings on two benchmarks with three model families, it is never significantly less accurate than the best fixed protocol and beats every fixed protocol on PeopleJoin. Reading who holds the data before deciding how to collaborate is a simple principle for agents that work over data no single agent holds.

\section*{Limitations}

We study two protocol families under a per-question communication budget, which we vary only on BIRD with Qwen. Correctness is a strict match of row multisets, complemented by row-level F1. Ownership in BIRD is assigned by a fixed hashing rule, which separates ownership from query shape by design, and PeopleJoin supplies simulated organizations whose members each hold at most one table. Organizations that keep related tables with one owner would produce more single-owner joins, the case in which owner cover and table count disagree. Such placements also change the mix of single- and two-owner questions, and Section~\ref{sec:main} reweights the measured accuracies to every such mix. A coordinator reaches at most six owners and thirteen tables per question, so very large catalogs may require retrieving candidate tables before drafting. The owner cover that defines the question strata is computed from reference SQL for analysis only, while OAR relies on a model-written draft whose quality depends on the model. OAR also assumes a public catalog of who holds which tables and one extra model call before any owner is contacted. Our experiments cover English questions over relational tables and three model families, and Relay+ runs with two of them. Owners that hold documents or other non-relational data, and budgets that differ across owners, are natural extensions.

\section*{Ethical Considerations}

All data come from two public research benchmarks, used as released: PeopleJoin-QA, built on Spider and distributed under CDLA-Permissive-2.0, and BIRD, distributed under CC BY-SA 4.0. Some BIRD databases contain public records about real people, such as the names and birth dates of athletes, and we report aggregate results and one illustrative trace that shows only an opaque record id. No human subjects were involved. Our setting is motivated by keeping data with their owners. In every protocol owners run queries locally and only result rows leave an owner. Under relay, some of these rows, typically join keys, also reach the next owner inside the coordinator's queries. Where one owner's keys must not reach another, routing can be restricted to fan-out, and our results measure what this costs on joins across owners. Private set intersection \citep{freedman2004psi} could instead let owners find matching keys without revealing the rest, at additional cost. Routing reads only the public catalog, but the catalog itself reveals which owner holds which data. A compromised coordinator could also send crafted queries to several owners. Our owners accept only read-only queries over their own tables and cap the rows they return. Deployments would also need access policies and auditing at each owner.

\section*{Acknowledgments}

AI assistants were used to implement the experiment pipeline, to produce the figures, and to draft and edit the text. Figure~\ref{fig:overview} was first generated with an AI tool and then revised and checked by hand. All reported numbers are computed by scripts from saved model outputs, and the authors verified the code, the numbers, the citations and the claims.

\bibliography{references}

\clearpage
\appendix

\section{Protocol Details}\label{app:protocols}

\paragraph{Contacted owners.} On PeopleJoin the coordinator acts for the asking user and ranks the other members by BM25 ($k_1=1.2$, $b=0.75$, stopwords removed) between the question and their public descriptions. The asking user and the top three members form the initial contacts, which fan-out uses as its owner set. Relay, Relay+ and the OAR draft may add at most two owners named in referral messages in the contacts' documents. On BIRD the four owners of each database are all contacted.

\paragraph{Fan-out.} Each contacted owner receives the question and its local schema with the instruction: ``Produce exactly one read-only SQLite SELECT or WITH SELECT query using only listed local tables and columns. Compute the most useful local partial result for the question; return JSON \{"sql":"..."\}. If the needed table is not here, return \{"sql":""\}. Do not assume access to other owners.'' Queries run locally with a limit of 100 rows and 8{,}000 bytes. The asking user's result is supplied to the receiver directly. The results of the other owners enter the receiver's input in contact order, each in full, as long as the serialized list fits the 2{,}048-token budget, and a result that would exceed it is left out. The receiver's instruction reads: ``Combine only the supplied local and peer results. Return JSON with unanswerable boolean and answer\_rows list of lists. If a needed relation was absent, do not fabricate it.''

\paragraph{Relay.} The planner sees the question, the owners' schemas, referral messages, prior results and the remaining budget. Its instruction reads: ``Plan the NEXT most informative read-only SQLite SELECT query on exactly one owner. For cross-owner joins, first obtain a small filtered set of join keys; on the next step use those literal keys in a query to another owner. Do not SELECT * or return full tables. Use WHERE/aggregate/LIMIT to send only relevant rows.'' The planner also sees the list of its earlier queries. It may issue five queries, including a second, different query to an owner it already queried. Each result is limited to 20 rows and 1{,}600 bytes. The executed SQL of every result is attached for the receiver after the last query, and 256 tokens of the budget are reserved for it. A run stops if this provenance does not fit. The receiver answers from executed, untruncated rows and abstains when needed rows or owners are absent.

\paragraph{Relay+.} Relay+ runs the same kind of planner loop with the same key-passing instruction. Each result carries its executed SQL from the moment it is returned, so the planner and the receiver both see it, and every query and result is charged against the budget as it is produced. A result that would exceed the remaining budget is replaced by a short status message, and no fixed share of the budget is reserved.

\paragraph{Planned parallel.} The planned variant of relay in Section~\ref{sec:mechanism} makes one planning call with the input of relay's first step. Its instruction reads: ``Before seeing any query results, plan at most 5 read-only SQLite SELECT queries, each for exactly one listed owner. Choose filtered, narrow partial results, including join-key columns and the attributes needed to answer cross-owner questions. Do not SELECT * or return full tables.'' The planned queries then run through relay's executor with its limits, budget, provenance and receiver. It was run on PJ-8b with DeepSeek and Sonnet and on BIRD with Qwen and DeepSeek.

\paragraph{OAR.} The draft prompt lists the tables and columns of the reachable owners and reads: ``Write one read-only SQLite SELECT or WITH SELECT over only these tables. The SQL result must contain precisely the requested answer columns and rows, with no explanatory columns. Return JSON \{"sql":"..."\}; return an empty sql string if no safe answer is possible.'' Tables are extracted from \texttt{FROM} and \texttt{JOIN} clauses with a regular expression, including those inside subqueries, and the owner cover of the extracted tables is computed exactly by enumerating owner subsets. Every table has a single holder except the row-split tables of PeopleJoin, which have two, so the cover equals the number of distinct holders unless a row-split table is involved, and the enumeration stays small. A name that no reachable owner holds, such as that of a common table expression, leaves the cover undefined and sends the question to fan-out. The route is stored before any protocol call. SQL errors are returned to the coordinator as a status, and a model reply that cannot be parsed ends the run, which then counts as incorrect.

\paragraph{Centralized reference.} The centralized reference executes the draft once on a merged database containing the tables of all reachable owners, with a limit of 100 rows and 8{,}000 bytes. It shares the draft with OAR and differs only in reading every owner's rows, which our setting forbids, so it measures the draft itself rather than a deployable protocol. It reaches 46.5, 51.3 and 42.9\% on PJ-7 with DeepSeek, Sonnet and Qwen, 71.7 and 73.7\% on PJ-Unseen with DeepSeek and Qwen, 44.9, 42.0 and 42.2\% on BIRD with the same three models, and 45.0 and 39.3\% on BIRD-Unseen with DeepSeek and Qwen.

\section{Evaluation Pools}\label{app:pools}

\paragraph{PeopleJoin.} PJ-8a contains 12 questions from each of eight organizations, selected by a hash of their public identifiers before any model was run. PJ-8b contains the remaining 494 questions of the same organizations. PJ-7 contains 511 questions from seven other organizations, locked by identifier before evaluation. PJ-Unseen draws 300 questions, by a hash of their identifiers, from the PeopleJoin-QA organizations that none of our earlier experiments used and whose member data are available. The sample covers 108 organizations, with at most three questions per organization and one question per duplicated reference SQL. Unanswerable questions are kept at their natural rate, 20.7\% of PJ-7 and 10.7\% of PJ-Unseen. Every PeopleJoin member holds at most one table. In 53 of the 123 organizations of our pools, the rows of one table are split between two members, which Section~\ref{sec:setup} treats as row-split.

\paragraph{BIRD.} Each database's tables are ordered by the SHA-256 hash of the string \texttt{v44-bird-20260928|} followed by the table name and dealt to $\min(4, n)$ owners in turn, where $n$ is the number of tables. The assignment depends only on table names, never on questions. Every owner holds a physically separate SQLite file with its own tables. BIRD uses the 500 Mini-Dev questions. BIRD-Unseen samples 300 development questions outside Mini-Dev with all 43 challenging, 150 moderate and 107 simple questions by the original difficulty labels, and it is scored against the November 2025 cleaned annotations.

\section{Additional Results}\label{app:results}

\subsection{The reversal (Section~\ref{sec:reversal})}\label{app:reversal}

\begin{table*}[t]
\centering\small
\setlength{\tabcolsep}{3.3pt}
\begin{tabular*}{\textwidth}{@{\extracolsep{\fill}}l rrr@{\hspace{9pt}}rrr@{\hspace{9pt}}rrr@{\hspace{9pt}}rrr}
\toprule
 & \multicolumn{3}{c}{Single-owner} & \multicolumn{3}{c}{Row-split} & \multicolumn{3}{c}{Two-owner} & \multicolumn{3}{c}{Unanswerable} \\
\cmidrule(lr){2-4}\cmidrule(lr){5-7}\cmidrule(lr){8-10}\cmidrule(lr){11-13}
Setting & FO & RL & OAR & FO & RL & OAR & FO & RL & OAR & FO & RL & OAR \\
\midrule
PJ-7, DeepSeek & 46.3 & 41.6 & 45.8 & 22.2 & 12.2 & 22.2 & 0.0 & 11.6 & 10.1 & 92.5 & 78.3 & 92.5 \\
PJ-7, Sonnet & 45.3 & 40.9 & 45.3 & 20.0 & 17.8 & 20.0 & 2.2 & 14.2 & 12.7 & 91.5 & 41.5 & 90.6 \\
PJ-7, Qwen & 51.2 & 45.8 & 50.7 & 22.2 & 13.3 & 22.2 & 0.0 & 12.7 & 11.9 & 91.5 & 90.6 & 91.5 \\
PJ-Unseen, DeepSeek & 77.6 & 67.3 & 76.2 & 55.2 & 10.3 & 51.7 & 1.4 & 47.9 & 47.9 & 75.0 & 58.3 & 66.7 \\
PJ-Unseen, Qwen & 80.3 & 74.1 & 80.3 & 69.0 & 24.1 & 69.0 & 1.4 & 38.0 & 38.0 & 75.0 & 72.2 & 72.2 \\
BIRD, DeepSeek & 47.3 & 28.8 & 45.4 & -- & -- & -- & 3.8 & 24.4 & 23.9 & -- & -- & -- \\
BIRD, Sonnet & 46.9 & 36.2 & 46.2 & -- & -- & -- & 2.7 & 28.3 & 24.6 & -- & -- & -- \\
BIRD, Qwen & 49.2 & 42.3 & 49.2 & -- & -- & -- & 3.1 & 28.7 & 26.3 & -- & -- & -- \\
BIRD-Unseen, DeepSeek & 54.4 & 36.8 & 51.5 & -- & -- & -- & 2.3 & 23.4 & 23.4 & -- & -- & -- \\
BIRD-Unseen, Qwen & 48.5 & 50.0 & 47.1 & -- & -- & -- & 2.9 & 33.9 & 33.3 & -- & -- & -- \\
\bottomrule
\end{tabular*}
\caption{Accuracy (\%) by reference owner cover. FO: fan-out, RL: relay. OAR follows fan-out on single-owner, row-split and unanswerable questions and relay on two-owner questions. BIRD has no row-split or unanswerable questions. On PJ-Unseen the last group also contains four questions whose reference tables could not be resolved.}
\label{tab:strata-main}
\end{table*}

\begin{table*}[t]
\centering\footnotesize
\setlength{\tabcolsep}{3.3pt}
\begin{tabular*}{\textwidth}{@{\extracolsep{\fill}}l rrrrr@{\hspace{12pt}}rrrrr@{\hspace{12pt}}rrrrr}
\toprule
 & \multicolumn{5}{c}{Single-owner} & \multicolumn{5}{c}{Two-owner} & \multicolumn{5}{c}{Three or more owners} \\
\cmidrule(lr){2-6}\cmidrule(lr){7-11}\cmidrule(lr){12-16}
Setting & FO & RL & RL+ & OAR & OAR+ & FO & RL & RL+ & OAR & OAR+ & FO & RL & RL+ & OAR & OAR+ \\
\midrule
PJ-7, DS & 46.3 & 41.6 & 41.9 & 45.8 & 46.1 & 0.0 & 11.6 & 8.2 & 10.1 & 7.1 & 0.0 & 0.0 & 0.0 & 0.0 & 0.0 \\
PJ-7, So & 45.3 & 40.9 & -- & 45.3 & -- & 2.2 & 14.2 & -- & 12.7 & -- & 0.0 & 21.7 & -- & 17.4 & -- \\
PJ-7, Qw & 51.2 & 45.8 & -- & 50.7 & -- & 0.0 & 12.7 & -- & 11.9 & -- & 0.0 & 0.0 & -- & 0.0 & -- \\
PJ-Unseen, DS & 77.6 & 67.3 & 68.7 & 76.2 & 75.5 & 1.4 & 47.9 & 40.8 & 47.9 & 40.8 & 0.0 & 29.4 & 17.6 & 29.4 & 17.6 \\
PJ-Unseen, Qw & 80.3 & 74.1 & 73.5 & 80.3 & 80.3 & 1.4 & 38.0 & 38.0 & 38.0 & 38.0 & 5.9 & 23.5 & 17.6 & 23.5 & 17.6 \\
BIRD, DS & 47.3 & 28.8 & 35.8 & 45.4 & 45.4 & 3.8 & 24.4 & 30.2 & 23.9 & 28.7 & 0.0 & 21.7 & 22.5 & 20.3 & 21.7 \\
BIRD, So & 46.9 & 36.2 & -- & 46.2 & -- & 2.7 & 28.3 & -- & 24.6 & -- & 0.0 & 20.3 & -- & 18.8 & -- \\
BIRD, Qw & 49.2 & 42.3 & 46.2 & 49.2 & 49.2 & 3.1 & 28.7 & 29.4 & 26.3 & 26.3 & 0.0 & 24.6 & 24.6 & 21.7 & 21.7 \\
BIRD-Unseen, DS & 54.4 & 36.8 & 44.1 & 51.5 & 51.5 & 2.3 & 23.4 & 31.0 & 23.4 & 29.8 & 2.4 & 17.1 & 19.5 & 19.5 & 22.0 \\
BIRD-Unseen, Qw & 48.5 & 50.0 & 52.9 & 47.1 & 47.1 & 2.9 & 33.9 & 32.2 & 33.3 & 32.2 & 2.4 & 17.1 & 22.0 & 17.1 & 22.0 \\
\bottomrule
\end{tabular*}
\caption{Accuracy (\%) by reference owner cover, including Relay+ and OAR+. FO: fan-out, RL: relay, RL+: Relay+. Single-owner questions exclude the row-split questions of Table~\ref{tab:strata-main}. Three-or-more groups hold 23 (PJ-7), 17 (PJ-Unseen), 69 (BIRD) and 41 (BIRD-Unseen) questions.}
\label{tab:strata}
\end{table*}

\begin{table}[t]
\centering\small
\setlength{\tabcolsep}{3pt}
\begin{tabular*}{\columnwidth}{@{\extracolsep{\fill}}lrr}
\toprule
Setting & Relay & Relay+ \\
\midrule
PJ-8a, DS & +24.0 [5.5, 43.1] & -- \\
PJ-8b, DS & +37.2 [28.1, 46.3] & -- \\
PJ-8b, So & +33.3 [23.5, 43.5] & -- \\
PJ-7, DS & +16.2 [8.7, 24.6] & +12.6 [5.6, 20.2] \\
PJ-7, So & +16.4 [8.0, 25.5] & -- \\
PJ-7, Qw & +18.1 [10.9, 25.8] & -- \\
PJ-Unseen, DS & +56.7 [41.9, 71.1] & +48.3 [33.7, 62.4] \\
PJ-Unseen, Qw & +42.7 [29.6, 56.6] & +43.4 [30.3, 56.8] \\
BIRD, DS & +39.1 [30.0, 48.1] & +38.0 [29.2, 46.8] \\
BIRD, So & +36.4 [26.7, 46.0] & -- \\
BIRD, Qw & +32.5 [24.7, 40.7] & +29.4 [21.2, 37.6] \\
BIRD-Unseen, DS & +38.7 [27.2, 50.7] & +38.9 [27.8, 50.7] \\
BIRD-Unseen, Qw & +29.5 [18.3, 41.0] & +24.8 [13.6, 35.7] \\
\bottomrule
\end{tabular*}
\caption{Interaction between protocol and ownership in points: the advantage of relay (or Relay+) over fan-out on two-owner questions minus its advantage on single-owner questions, with 95\% intervals.}
\label{tab:interaction}
\end{table}

\paragraph{Per-stratum accuracy.} Table~\ref{tab:strata} extends Table~\ref{tab:strata-main} with Relay+, OAR+ and the questions that need three or more owners, where fan-out answers three questions in total across all settings. On the row-split questions of PJ-8a (2) and PJ-8b (14), fan-out answers 0.0\% on PJ-8a and 0.0 and 28.6\% on PJ-8b with DeepSeek and Sonnet, against 25.0, 7.1 and 14.3\% for relay. On the unanswerable PeopleJoin questions with DeepSeek, Relay+ abstains correctly on 88.7\% (PJ-7) and 66.7\% (PJ-Unseen) and OAR+ on 92.9\% and 72.2\%. On BIRD, eight Mini-Dev and twenty BIRD-Unseen references could not be parsed into tables and are omitted from the strata.

\paragraph{Interaction.} Table~\ref{tab:interaction} lists the interaction for every setting. Per database, the interaction is positive in 7 of 7 PJ-7 databases with DeepSeek and Qwen and 6 of 7 with Sonnet, in all 11 BIRD databases with each of the three models, in 8 of 8 and 7 of 8 PJ-8b databases with DeepSeek and Sonnet, and in 6 of the 8 PJ-8a databases, whose strata contain between one and nine questions each.

\paragraph{Ownership with the query shape fixed.} The 95\% intervals of the differences in Table~\ref{tab:natural} are $[19.4, 47.1]$, $[16.7, 44.4]$, $[26.6, 47.5]$, $[25.7, 66.5]$ and $[14.2, 48.7]$.

\subsection{The mechanism (Section~\ref{sec:mechanism})}\label{app:mechanism}

\begin{table*}[t]
\centering\small
\setlength{\tabcolsep}{5pt}
\begin{tabular*}{\textwidth}{@{\extracolsep{\fill}}l rrrr@{\hspace{16pt}}rrrrr}
\toprule
 & \multicolumn{4}{c}{Fan-out} & \multicolumn{5}{c}{Relay} \\
\cmidrule(lr){2-5}\cmidrule(lr){6-10}
Setting & Correct & Abst. & Wrong & Stop. & Correct & Abst. & Key & No key & Stop. \\
\midrule
PJ-7, DeepSeek & 0.0 & 80.6 & 19.4 & 0.0 & 11.9 & 42.5 & 23.1 & 16.4 & 6.0 \\
PJ-7, Qwen & 0.0 & 81.3 & 17.9 & 0.7 & 12.7 & 54.5 & 20.1 & 12.7 & 0.0 \\
PJ-Unseen, DeepSeek & 1.4 & 90.1 & 8.5 & 0.0 & 47.9 & 19.7 & 18.3 & 8.5 & 5.6 \\
BIRD, DeepSeek & 4.1 & 78.2 & 16.4 & 1.4 & 25.3 & 18.4 & 28.7 & 4.8 & 22.9 \\
BIRD, Qwen & 3.1 & 77.8 & 18.1 & 1.0 & 28.7 & 27.0 & 32.1 & 4.8 & 7.5 \\
BIRD-Unseen, DeepSeek & 2.3 & 80.1 & 14.6 & 2.9 & 23.4 & 13.5 & 28.7 & 4.7 & 29.8 \\
\bottomrule
\end{tabular*}
\caption{Outcomes on two-owner questions (\% of questions, first run). Abst.: the receiver declares the question unanswerable or returns no rows. Wrong: wrong rows. Key and No key: wrong rows after relay passed a key or without one. Stop.: the run ends early, for relay mostly because its provenance reserve is exhausted.}
\label{tab:failures}
\end{table*}

\begin{table}[t]
\centering\footnotesize
\begin{tabular}{@{}>{\raggedright\arraybackslash}p{\columnwidth}@{}}
\toprule
\textbf{Q:} What is the amount of the funds that the Vice President received? (BIRD, \texttt{student\_club}) \\
\midrule
\textit{Step 1}, owner of \texttt{member}: \texttt{SELECT member\_id FROM member WHERE position = \textquotesingle Vice President\textquotesingle{} LIMIT 5} $\rightarrow$ \texttt{recD078PnS3x2doBe} \\[2pt]
\textit{Step 2}, owner of \texttt{income}: \texttt{SELECT amount FROM income WHERE link\_to\_member = \textquotesingle recD078PnS3x2doBe\textquotesingle{} LIMIT 5} $\rightarrow$ \texttt{50} \\[2pt]
Relay answers 50, which is correct. Under fan-out each owner answers alone, and the receiver abstains. \\
\bottomrule
\end{tabular}
\caption{A relay run with DeepSeek in which the member id returned by one owner becomes the filter of the query to the next owner.}
\label{tab:example}
\end{table}

\begin{table}[t]
\centering\small
\setlength{\tabcolsep}{2.5pt}
\begin{tabular*}{\columnwidth}{@{\extracolsep{\fill}}lrrrrr}
\toprule
 & & & \multicolumn{2}{c}{Accuracy (\%)} & \\
\cmidrule(lr){4-5}
Setting & Runs & Key (\%) & key & no key & Share (\%) \\
\midrule
PJ-7, DS & 268 & 52.6 & 12.1 & 11.0 & 54.8 \\
PJ-7, So & 134 & 65.7 & 15.9 & 10.9 & 73.7 \\
PJ-7, Qw & 134 & 44.8 & 15.0 & 10.8 & 52.9 \\
PJ-Uns., DS & 71 & 74.6 & 58.5 & 16.7 & 91.2 \\
PJ-Uns., Qw & 71 & 64.8 & 50.0 & 16.0 & 85.2 \\
BIRD, DS & 586 & 81.2 & 28.8 & 5.5 & 95.8 \\
BIRD, So & 293 & 91.5 & 28.7 & 24.0 & 92.8 \\
BIRD, Qw & 293 & 74.4 & 35.8 & 8.0 & 92.9 \\
BIRD-Uns., DS & 171 & 81.9 & 27.9 & 3.2 & 97.5 \\
BIRD-Uns., Qw & 171 & 84.8 & 37.9 & 11.5 & 94.8 \\
\bottomrule
\end{tabular*}
\caption{Key passing in relay runs on two-owner questions. Key: share of runs in which a later query contains a literal value that a different owner returned earlier. Share: fraction of the correct runs that passed a key.}
\label{tab:keys}
\end{table}

\begin{table}[t]
\centering\small
\setlength{\tabcolsep}{3.2pt}
\begin{tabular*}{\columnwidth}{@{\extracolsep{\fill}}lrrrrrr}
\toprule
 & \multicolumn{3}{c}{At most 20 rows} & \multicolumn{3}{c}{More than 20} \\
\cmidrule(lr){2-4}\cmidrule(lr){5-7}
Setting & Qs & FO & RL & Qs & FO & RL \\
\midrule
PJ-8a, DS & 34 & 61.8 & 55.9 & 7 & 14.3 & 0.0 \\
PJ-8b, DS & 218 & 59.9 & 49.5 & 57 & 33.3 & 0.0 \\
PJ-8b, So & 218 & 52.8 & 55.0 & 57 & 21.1 & 1.8 \\
PJ-7, DS & 173 & 48.6 & 48.8 & 30 & 33.3 & 0.0 \\
PJ-7, So & 173 & 46.8 & 47.4 & 30 & 36.7 & 3.3 \\
PJ-7, Qw & 173 & 53.8 & 53.8 & 30 & 36.7 & 0.0 \\
PJ-Uns., DS & 147 & 77.6 & 67.3 & 0 & -- & -- \\
PJ-Uns., Qw & 147 & 80.3 & 74.1 & 0 & -- & -- \\
BIRD, DS & 118 & 52.1 & 31.8 & 12 & 0.0 & 0.0 \\
BIRD, So & 118 & 51.7 & 39.8 & 12 & 0.0 & 0.0 \\
BIRD, Qw & 118 & 54.2 & 46.6 & 12 & 0.0 & 0.0 \\
BIRD-Uns., DS & 64 & 57.8 & 39.1 & 4 & 0.0 & 0.0 \\
BIRD-Uns., Qw & 64 & 51.6 & 53.1 & 4 & 0.0 & 0.0 \\
\bottomrule
\end{tabular*}
\caption{Single-owner accuracy (\%) by the number of rows in the reference answer. Relay limits each step to 20 rows.}
\label{tab:size}
\end{table}

\paragraph{How runs fail.} Table~\ref{tab:failures} breaks down the outcomes on two-owner questions. Fan-out rarely returns a wrong join. Its receiver abstains on 78--90\% of the questions, because the rows that a join needs never arrive. Relay's errors split between abstentions, wrong rows after a key was passed, and runs stopped by the fixed provenance reserve, which Relay+ removes.

\paragraph{Row visibility.} In the PJ-8b ablation of Section~\ref{sec:mechanism}, showing the planner the returned rows changes single-owner accuracy by $-0.4$ points ($[-2.7, 2.0]$) and two-owner accuracy by $+18.1$ points ($[12.0, 24.6]$). Adding the key-passing instruction to the planner that sees the rows moves two-owner accuracy from 23.2\% to 24.3\%. Restricting the two-owner questions to those whose reference SQL contains an explicit JOIN (126 questions) gives $+18.7$ points ($[12.3, 25.4]$), and the effect is positive in 7 of the 8 organizations. On a 186-question subset with Sonnet, showing the rows raises two-owner accuracy from 16.3\% to 22.4\% ($+6.1$ points, $[0.0, 14.3]$) and leaves single-owner accuracy at 55.0\% versus 56.0\%. On BIRD with Qwen, it raises two-owner accuracy from 2.7\% to 20.5\% ($+17.7$ points, $[13.7, 22.2]$) and changes single-owner accuracy by $-0.8$ points ($[-3.8, 2.3]$), with the same number of calls in both conditions. There, the explicit key-passing instruction adds $+8.9$ points on two-owner questions ($[4.4, 13.3]$). With DeepSeek on BIRD, showing the rows raises two-owner accuracy from 3.4\% to 19.3\% ($+15.9$ points, $[11.7, 20.3]$) and changes single-owner accuracy by $+3.1$ points ($[-3.1, 9.2]$), and the key-passing instruction adds $+11.2$ points ($[7.2, 15.3]$). Three BIRD questions whose relay call ended without a reply are excluded from these DeepSeek contrasts, since such calls are never re-sent.

\paragraph{Key passing in Relay+.} Relay+ passes a key in 80.7\% (BIRD, DeepSeek), 74.4\% (BIRD, Qwen), 78.9\% (BIRD-Unseen, Qwen), 69.0\% (PJ-Unseen, Qwen) and 52.6\% (PJ-7, DeepSeek) of its two-owner runs, and these runs produce 94.4\%, 89.5\%, 96.4\%, 85.2\% and 72.7\% of its correct two-owner answers. The literal matching behind Table~\ref{tab:keys} counts a returned string only when it appears quoted in a later query, and a returned number only when it appears as a separate token that is absent from the question and is not the argument of \texttt{LIMIT} or \texttt{OFFSET}.

\paragraph{Planned parallel.} On PJ-8b, relay exceeds the planned variant on two-owner questions by 18.1 points with DeepSeek ($[10.9, 25.4]$) and 7.2 points with Sonnet ($[0.0, 14.5]$). Fan-out exceeds it on single-owner questions by 22.9 ($[17.1, 28.9]$) and 14.2 points ($[8.7, 19.6]$). Overall the planned variant reaches 25.4\% and 30.4\%, below both fan-out and relay. On BIRD with Qwen, relay exceeds it on two-owner questions by 24.6 points ($[19.5, 29.7]$), while it differs from fan-out there by only $+1.0$ point ($[-1.7, 4.1]$), and fan-out exceeds it on single-owner questions by 21.5 points ($[13.8, 30.0]$). Overall it reaches 10.0\% there. With DeepSeek on BIRD, relay exceeds it on two-owner questions by 22.1 points ($[17.1, 27.2]$), fan-out differs from it there by $+1.4$ points ($[-0.7, 3.4]$) and exceeds it on single-owner questions by 35.0 points ($[26.9, 43.5]$), and it reaches 4.6\% overall, with one question excluded for the same reason as above.

\paragraph{Answer size and reach.} Table~\ref{tab:size} splits single-owner questions by the size of the reference answer. Restricted to questions whose reference answer has at most 20 rows, the interaction is $+24.6$ points ($[3.4, 46.4]$) on PJ-8a, $+35.3$ ($[25.5, 45.3]$) and $+33.0$ ($[21.9, 43.8]$) on PJ-8b with DeepSeek and Sonnet, $+14.6$ ($[6.0, 24.0]$), $+14.8$ ($[5.2, 24.8]$) and $+16.3$ ($[8.1, 25.0]$) on PJ-7 with DeepSeek, Sonnet and Qwen, $+55.9$ ($[41.6, 70.2]$) and $+41.8$ ($[29.1, 55.1]$) on PJ-Unseen with DeepSeek and Qwen, $+43.7$ ($[33.7, 53.8]$), $+41.0$ ($[30.3, 51.9]$) and $+36.8$ ($[28.1, 45.8]$) on BIRD with DeepSeek, Sonnet and Qwen, and $+43.4$ ($[30.9, 56.3]$) and $+34.7$ ($[22.3, 47.0]$) on BIRD-Unseen with DeepSeek and Qwen. On PeopleJoin two-owner questions whose two owners are both among fan-out's initial contacts (58 on PJ-7, 54 on PJ-Unseen), fan-out answers 0.0, 5.2 and 0.0\% on PJ-7 with DeepSeek, Sonnet and Qwen and 1.9\% on PJ-Unseen with both models, against 13.8, 15.5, 10.3, 55.6 and 46.3\% for relay.

\paragraph{Communication budget.} On BIRD with Qwen we rerun fan-out and relay with $B=1{,}024$ and $B=8{,}192$ tokens. Fan-out answers 49.2\% of the single-owner and 3.1\% of the two-owner questions at every budget, because its peer results exceed 1{,}024 tokens in only 9 of the 500 questions. Relay answers 35.4, 42.3 and 43.8\% of the single-owner and 23.5, 28.7 and 29.4\% of the two-owner questions at $B=1{,}024$, 2{,}048 and 8{,}192, and it stops early in 21.6\% of its runs at the smallest budget against 5.6\% at the largest. The interaction is $+34.3$ ($[25.9, 43.1]$), $+32.5$ ($[24.7, 40.7]$) and $+31.7$ points ($[23.7, 39.9]$). OAR differs from relay by $+1.8$, $-0.2$ and $-0.8$ points, with every interval including zero, and exceeds fan-out by 13.8 to 17.0 points.

\subsection{Routing (Section~\ref{sec:oar})}\label{app:routing}

\begin{table*}[t]
\centering\small
\setlength{\tabcolsep}{4pt}
\begin{tabular*}{\textwidth}{@{\extracolsep{\fill}}lrrrrr}
\toprule
Setting & OAR$-$FO & OAR$-$RL & OAR$-$best & OAR+$-$RL+ & OAR+$-$OAR \\
\midrule
PJ-7, DeepSeek & +2.4 [1.0, 4.0] & +5.1 [2.3, 7.9] & +2.4 [1.0, 4.0] & +3.3 [0.9, 5.9] & $-$0.6 [$-$1.5, 0.3] \\
PJ-7, Sonnet & +3.3 [1.4, 5.3] & +11.5 [7.8, 15.3] & +3.3 [1.4, 5.3] & -- & -- \\
PJ-7, Qwen & +2.9 [1.4, 4.5] & +2.7 [0.6, 5.1] & +2.7 [0.6, 5.1] & -- & -- \\
PJ-Unseen, DeepSeek & +10.7 [6.3, 15.0] & +9.3 [5.0, 13.7] & +9.3 [5.0, 13.7] & +8.3 [4.0, 12.7] & $-$2.0 [$-$4.7, 0.3] \\
PJ-Unseen, Qwen & +9.3 [5.7, 13.0] & +7.3 [3.0, 11.7] & +7.3 [3.0, 11.7] & +8.3 [4.0, 12.7] & +0.0 [$-$2.0, 2.0] \\
BIRD, DeepSeek & +14.3 [11.1, 17.6] & +3.6 [1.4, 5.9] & $-$1.5 [$-$4.4, 1.4] & +1.3 [$-$0.7, 3.3] & +2.8 [0.7, 4.9] \\
BIRD, Sonnet & +15.4 [12.2, 18.6] & +0.0 [$-$2.6, 2.6] & +0.0 [$-$2.6, 2.6] & -- & -- \\
BIRD, Qwen & +16.8 [13.6, 20.2] & $-$0.2 [$-$2.2, 1.8] & $-$1.6 [$-$4.4, 1.2] & $-$1.6 [$-$3.8, 0.4] & +0.0 [$-$1.8, 1.8] \\
BIRD-Unseen, DeepSeek & +14.0 [10.0, 18.3] & +3.7 [1.3, 6.3] & $-$2.7 [$-$6.3, 1.0] & +1.3 [$-$1.0, 3.7] & +4.0 [1.3, 7.0] \\
BIRD-Unseen, Qwen & +19.0 [14.3, 23.7] & $-$1.0 [$-$3.3, 1.0] & $-$1.3 [$-$4.7, 2.0] & $-$1.3 [$-$3.3, 0.7] & +0.0 [$-$2.7, 2.7] \\
\bottomrule
\end{tabular*}
\caption{Paired differences in points with 95\% intervals. FO: fan-out, RL: relay, RL+: Relay+. ``Best'' is the most accurate fixed protocol of the setting, including Relay+ where it runs.}
\label{tab:contrasts}
\end{table*}

\begin{figure}[t]
\centering
\includegraphics[width=\columnwidth]{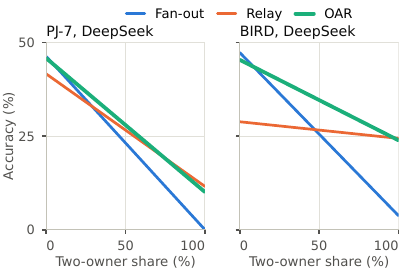}
\caption{Expected accuracy as the share of two-owner questions varies, obtained by reweighting the measured single-owner and two-owner accuracies. OAR stays near the better of fan-out and relay for every mix and exceeds both at intermediate mixes.}
\label{fig:mix}
\end{figure}

\begin{table*}[t]
\centering\small
\setlength{\tabcolsep}{4pt}
\begin{tabular*}{\textwidth}{@{\extracolsep{\fill}}l rrr@{\hspace{14pt}}rrr@{\hspace{14pt}}rrr}
\toprule
 & \multicolumn{3}{c}{Single-owner} & \multicolumn{3}{c}{Two-owner} & \multicolumn{3}{c}{All questions} \\
\cmidrule(lr){2-4}\cmidrule(lr){5-7}\cmidrule(lr){8-10}
Setting & FO & RL & OAR & FO & RL & OAR & FO & RL & OAR \\
\midrule
PJ-7, DeepSeek & 50.8 & 47.1 & 49.5 & 1.1 & 13.1 & 12.4 & 43.0 & 40.9 & \textbf{45.7} \\
PJ-7, Sonnet & 49.5 & 42.4 & 49.1 & 3.2 & 18.7 & 17.7 & 42.1 & 33.6 & \textbf{46.4} \\
PJ-7, Qwen & 53.1 & 51.2 & 52.6 & 1.2 & 15.2 & 15.1 & 43.3 & 45.9 & \textbf{47.1} \\
PJ-Unseen, DeepSeek & 80.0 & 69.0 & 78.6 & 3.3 & 54.7 & 52.5 & 54.9 & 59.8 & \textbf{67.4} \\
PJ-Unseen, Qwen & 81.0 & 74.9 & 81.0 & 3.3 & 45.6 & 44.6 & 57.4 & 62.7 & \textbf{68.2} \\
BIRD, DeepSeek & 50.3 & 31.0 & 48.2 & 4.2 & 26.8 & 26.3 & 15.6 & 27.5 & \textbf{31.2} \\
BIRD, Sonnet & 49.6 & 36.8 & 48.5 & 3.3 & 31.2 & 27.3 & 14.9 & 31.3 & \textbf{31.6} \\
BIRD, Qwen & 52.0 & 46.0 & 52.8 & 3.6 & 33.1 & 30.7 & 15.6 & \textbf{35.4} & 35.1 \\
BIRD-Unseen, DeepSeek & 55.7 & 36.9 & 52.4 & 2.5 & 25.5 & 25.4 & 14.4 & 25.6 & \textbf{29.4} \\
BIRD-Unseen, Qwen & 49.9 & 50.8 & 47.9 & 3.8 & 36.7 & 35.6 & 13.8 & \textbf{35.0} & 33.7 \\
\bottomrule
\end{tabular*}
\caption{Row-level F1 (\%), which gives partial credit for overlapping rows. FO: fan-out, RL: relay. Bold marks the highest F1 over all questions in each row.}
\label{tab:f1}
\end{table*}

\begin{table}[t]
\centering\small
\setlength{\tabcolsep}{3.0pt}
\begin{tabular*}{\columnwidth}{@{\extracolsep{\fill}}lrrrrrr}
\toprule
 & \multicolumn{2}{c}{Draft} & \multicolumn{4}{c}{Accuracy (\%)} \\
\cmidrule(lr){2-3}\cmidrule(lr){4-7}
Setting & Prec. & Rec. & Oracle & OAR & Table & JOIN \\
\midrule
PJ-7, DS & 94.9 & 65.6 & 42.6 & 42.0 & 41.1 & 40.6 \\
PJ-7, So & 96.9 & 59.2 & 43.4 & 42.7 & 43.1 & 42.9 \\
PJ-7, Qw & 94.5 & 66.2 & 44.6 & 44.2 & 43.8 & 43.8 \\
PJ-Uns., DS & 87.9 & 90.9 & 65.3 & 63.3 & -- & -- \\
PJ-Uns., Qw & 93.7 & 84.1 & 65.3 & 65.0 & -- & -- \\
BIRD, DS & 97.1 & 86.7 & 29.6 & 28.8 & 28.6 & 28.4 \\
BIRD, So & 97.5 & 74.3 & 31.6 & 29.2 & 29.4 & 29.2 \\
BIRD, Qw & 99.0 & 83.1 & 33.0 & 31.4 & 31.4 & 31.0 \\
BIRD-Uns., DS & 96.9 & 88.2 & 28.0 & 28.0 & -- & -- \\
BIRD-Uns., Qw & 97.3 & 85.4 & 32.7 & 32.0 & -- & -- \\
\bottomrule
\end{tabular*}
\caption{How the draft routes. Precision and recall of the draft's multi-owner ($\kappa\geq 2$) prediction against the reference cover. Oracle routes by the reference cover with the same relay branch as OAR. Table and JOIN route the same draft on its table count or on the presence of a JOIN.}
\label{tab:analysis}
\end{table}

\begin{table}[t]
\centering\footnotesize
\setlength{\tabcolsep}{1.5pt}
\begin{tabular*}{\columnwidth}{@{\extracolsep{\fill}}lrrrrr}
\toprule
Setting & Best & OAR & 8B & $\Delta$ [95\% CI] & Rec. \\
\midrule
PJ-7, DS & 39.5 & 42.0 & 39.6 & +0.1 [$-$1.8, 2.0] & 61.1 \\
PJ-7, So & 39.3 & 42.7 & 38.6 & $-$0.8 [$-$3.3, 1.6] & 61.1 \\
PJ-7, Qw & 41.5 & 44.2 & 43.6 & +2.2 [0.0, 4.3] & 61.1 \\
PJ-Uns., DS & 54.0 & 63.3 & 60.3 & +6.3 [2.0, 11.0] & 78.4 \\
PJ-Uns., Qw & 57.7 & 65.0 & 60.7 & +3.0 [$-$1.3, 7.3] & 78.4 \\
BIRD, DS & 30.3 & 28.8 & 26.5 & $-$3.8 [$-$6.8, $-$0.7] & 67.7 \\
BIRD, So & 29.2 & 29.2 & 28.4 & $-$0.8 [$-$3.4, 1.8] & 67.7 \\
BIRD, Qw & 33.0 & 31.4 & 29.6 & $-$3.4 [$-$6.2, $-$0.6] & 67.7 \\
BIRD-Uns., DS & 30.7 & 28.0 & 24.3 & $-$6.3 [$-$10.7, $-$2.0] & 67.5 \\
BIRD-Uns., Qw & 33.3 & 32.0 & 26.7 & $-$6.7 [$-$10.7, $-$2.7] & 67.5 \\
\bottomrule
\end{tabular*}
\caption{OAR with drafts from Llama-3.1-8B (8B), with the executors of each setting. Best: the most accurate fixed protocol, as in Table~\ref{tab:main}. OAR: drafts from the executing model. $\Delta$: 8B minus Best in points. Rec.: the share of multi-owner questions that the 8B draft recognizes, the same for every executor of a pool.}
\label{tab:weak}
\end{table}

\paragraph{Paired contrasts.} Table~\ref{tab:contrasts} lists the paired contrasts behind Section~\ref{sec:main}.

\paragraph{Partial credit.} Table~\ref{tab:f1} replaces exact multiset match with row-level F1, the harmonic mean of row precision and recall under the same value normalization, with correct abstention scored as 1. Row-level F1 preserves the pattern of exact match. Fan-out earns at most 4.2 F1 on two-owner questions, so its failure there is not a near miss, and OAR has the highest F1 over all questions in eight of the ten settings.

\paragraph{Unanswerable questions.} On the unanswerable PJ-7 questions, fan-out abstains correctly in 91.5--92.5\% of the runs, since owners whose tables lack the requested relation decline to write a query. Relay answers from partial rows in 8--25\% of the runs. OAR sends most unanswerable questions to fan-out. With Sonnet, relay's receiver often adds an explanation field when it abstains, and the strict format check counts these replies as wrong, including on 43 of the 106 unanswerable PJ-7 questions. Counting them as abstentions raises relay from 31.1 to 39.5\% on PJ-7 with Sonnet, and OAR still leads every fixed protocol, by 3.1 points (95\% CI $[0.2, 6.1]$).

\paragraph{Terminated relay runs.} With the fixed 256-token provenance reserve, 103 of 1{,}000 DeepSeek runs, 88 of 500 Sonnet runs and 27 of 500 Qwen runs of relay on BIRD stop because the provenance of an executed query no longer fits. Relay+ has no such stops, and its remaining terminations affect 28 of 1{,}000 runs with DeepSeek and 4 of 500 with Qwen. Counting every cause, relay stops early in 213 of its 1{,}000 DeepSeek runs and 30 of its 500 Qwen runs on BIRD.

\paragraph{Cost.} Relative to relay, fan-out costs 0.74 and 0.63 times as much on PJ-7 with DeepSeek and Sonnet, and 0.62 and 0.71 times on BIRD. OAR costs 1.07, 0.89, 1.18 and 1.08 times as much as relay in the same four settings, including its draft call, and 1.45, 1.41, 1.90 and 1.52 times as much as fan-out. Qwen runs locally and has no API cost. Per question, fan-out makes 3.9 to 4.8 model calls, relay 3.4 to 4.6 and OAR 5.1 to 5.8, including its draft. With fan-out's owners queried one after another, the measured time per question on PJ-7 is 3.1, 15.1 and 5.9 seconds for fan-out with DeepSeek, Sonnet and Qwen, against 2.9, 17.4 and 6.9 for relay and 3.9, 19.5 and 8.8 for OAR. On BIRD it is 3.7, 23.4 and 6.4 seconds for fan-out, 3.7, 22.9 and 10.6 for relay and 4.8, 28.1 and 13.5 for OAR. With Qwen on PJ-Unseen and BIRD-Unseen it is 5.1 and 6.3 seconds for fan-out, 7.9 and 11.4 for relay and 8.8 and 14.1 for OAR. The DeepSeek runs on the unseen pools did not record time.

\paragraph{Where the draft misroutes.} Most routing errors send a multi-owner question to fan-out rather than the reverse. Against the reference cover, first-run drafts flag a single-owner question as multi-owner 3 to 11 times per setting. A multi-owner question is missed when the draft names a table that is not in the catalog (3 to 63 times per setting), when it joins fewer owners than the reference (4 to 26), or when it is empty (up to 52, mostly with Sonnet on PJ-7). On BIRD, the unlisted name is almost always a common table expression that the draft defines itself. Dropping such names, as a parser that resolves these expressions would, sends 151 of the 159 affected BIRD questions to relay. Re-routing the saved runs with such a parser would raise OAR's accuracy by 0.0 to 1.0 points in the five BIRD settings. All reported results keep the frozen parser. On PeopleJoin, the unlisted name is mostly a plausible table that the catalog does not list.

\paragraph{Owner cover against table count.} On the 44 BIRD questions that join two tables of one owner, OAR minus table-count routing is $+6.8$ ($[-4.5, 18.2]$), $+4.5$ ($[-6.8, 18.2]$) and $+9.1$ points ($[2.3, 18.2]$) with DeepSeek, Sonnet and Qwen.

\paragraph{Other routing signals.} Table~\ref{tab:main} compares OAR with routing on other signals. Self and Ask make one call per question with Qwen or DeepSeek and receive the same owner catalog as the draft. Self chooses fan-out or relay from their descriptions, Ask sends a question to relay when the model names two or more owners, and difficulty routing sends the questions that BIRD labels simple to fan-out and the rest to relay. Self is told that under fan-out ``each owner receives the question with only its own tables, writes one query and returns its result,'' while under relay ``a planner sees all owners' tables and queries one owner at a time, and each later query can use the rows returned by earlier queries.'' OAR minus Self is $+2.0$ ($[0.6, 3.5]$), $+2.3$ ($[0.0, 5.0]$), $+4.0$ ($[1.8, 6.2]$) and $+1.3$ points ($[-1.3, 4.0]$) on PJ-7, PJ-Unseen, BIRD and BIRD-Unseen. OAR minus Ask is $+0.8$ ($[0.0, 1.8]$), $+5.0$ ($[2.3, 8.0]$), $0.0$ ($[-1.0, 1.0]$) and $-0.3$ points ($[-1.3, 0.7]$). Ask sends 46.7\% and 89.7\% of the row-split questions of PJ-7 and PJ-Unseen to relay, against 0.0\% and 6.9\% for OAR, and Self sends 20--29\% of the single-owner questions there. With DeepSeek, OAR minus Self is $+3.5$ ($[2.0, 5.3]$), $+2.3$ ($[0.0, 4.7]$), $+1.4$ ($[-0.3, 3.2]$) and $+2.3$ points ($[0.3, 4.7]$) in the same four pools, OAR minus Ask is $+2.2$ ($[0.7, 3.7]$), $+5.0$ ($[2.7, 7.7]$), $+0.3$ ($[-0.9, 1.6]$) and $+0.7$ points ($[-0.7, 2.0]$), and Ask sends 71.1\% and 89.7\% of the row-split questions to relay, against 0.0\% and 3.4\% for OAR. Four of the 3{,}222 DeepSeek routing calls ended without a reply and their questions are excluded. OAR minus difficulty routing is $+8.0$ ($[5.5, 10.7]$), $+7.0$ ($[4.2, 10.0]$), $+7.8$ ($[5.2, 10.6]$), $+8.0$ ($[5.0, 11.3]$) and $+6.0$ points ($[3.0, 9.3]$) in the five BIRD settings, where 30\% of the BIRD questions and 32\% of the BIRD-Unseen questions are labeled simple. For BIRD-Unseen these labels come from the cleaned annotations we score against, while the sample was drawn by the original labels (Appendix~\ref{app:pools}).

\paragraph{Retry after fan-out.} We simulate a hybrid from the saved runs of each question: fan-out, followed by relay's run whenever fan-out's receiver abstains or its run stops. Used alone, the hybrid differs from OAR by $-1.9$ ($[-3.8, 0.1]$), $-8.2$ ($[-11.4, -5.1]$) and $-0.4$ points ($[-1.6, 0.8]$) on PJ-7 with DeepSeek, Sonnet and Qwen, by $0.0$ ($[-3.0, 3.0]$) and $+1.0$ ($[-1.0, 3.3]$) on PJ-Unseen with DeepSeek and Qwen, by $+0.1$ ($[-1.6, 1.8]$), $+2.2$ ($[0.8, 3.8]$) and $-2.2$ ($[-4.2, -0.2]$) on BIRD with DeepSeek, Sonnet and Qwen, and by $+0.7$ ($[-1.0, 2.3]$) and $+0.7$ ($[-1.3, 2.7]$) on BIRD-Unseen. Applied after OAR's route, the retry changes accuracy by $-1.3$ ($[-3.0, 0.5]$), $-8.2$ ($[-11.2, -5.3]$), $+0.4$ ($[-0.6, 1.4]$), $0.0$ ($[-2.3, 2.3]$) and $+1.3$ points ($[-0.7, 3.3]$) in the same PeopleJoin settings, and by $+0.8$ ($[0.2, 1.6]$), $+2.2$ ($[1.0, 3.6]$), $+1.4$ ($[0.4, 2.4]$), $+0.7$ ($[0.0, 1.7]$) and $+1.3$ points ($[0.3, 2.7]$) on BIRD and BIRD-Unseen. On PJ-7 its loss comes from unanswerable questions, which contribute $-2.9$, $-10.2$ and $-0.4$ points with DeepSeek, Sonnet and Qwen, while single-owner questions contribute $+1.5$, $+1.6$ and $+0.8$. Two-owner questions that OAR sends to fan-out are answered 0.0--10.0\% of the time, and those it sends to relay 15.4--52.3\%.

\paragraph{A smaller drafter.} Table~\ref{tab:weak} routes the saved fan-out and relay runs with drafts written by Llama-3.1-8B-Instruct with 5-bit weights, under the same prompt and parser as OAR. The smaller model names a table that is not in the catalog in 19--30\% of its drafts, which sends the question to fan-out. It recognizes 61--78\% of the multi-owner questions, and 69\% of its multi-owner predictions are correct on PeopleJoin and 90--92\% on BIRD. PeopleJoin tolerates these misses because most of its questions are best answered by fan-out. On BIRD, where most questions need several owners, each miss gives up relay's accuracy on that question, yet OAR with the 8B drafter remains 10 to 15 points above fan-out in every setting.

\section{Statistical Procedures}\label{app:stats}

For a contrast between two protocols we average each protocol's score over runs within a question, take the per-question difference, and report its mean with a 95\% percentile interval from 10{,}000 bootstrap resamples of questions, drawn with seed 20260927. For the interaction we compute each question's relay-minus-fan-out difference, resample the single-owner and two-owner strata separately 10{,}000 times with seed 20260928, and take the difference of the stratum means. All intervals are recomputed from the per-question outcomes with these procedures. Those in Table~\ref{tab:contrasts} agree to within 0.4 points with the intervals of the pre-specified analyses. The interaction intervals in Table~\ref{tab:interaction} use the strata of Section~\ref{sec:setup}, which set row-split questions apart. On these strata, a stratum-centered bootstrap test of the interaction with Holm correction across the six PJ-7 and BIRD settings gives adjusted $p<0.001$ in all of them. The pre-specified paired sign-flip tests with Holm correction give OAR's advantage over fan-out an adjusted $p\leq 0.004$ in the same six settings. Per-database estimates are descriptive and treat each database as one unit. Questions over one database share its schema and need not be independent. Resampling whole databases, which are organizations on PeopleJoin, instead of questions keeps every interaction interval above zero, with lower bounds from $+5.9$ to $+42.3$ points. The intervals of OAR minus the best fixed protocol become $[0.9, 4.7]$, $[2.0, 4.9]$ and $[0.0, 6.2]$ on PJ-7 with DeepSeek, Sonnet and Qwen and $[4.7, 13.8]$ and $[3.0, 11.8]$ on PJ-Unseen with DeepSeek and Qwen. Counted per organization and model, OAR is at least as accurate as the best fixed protocol in 19 of 21 cases on PJ-7 and in 198 of 216 on PJ-Unseen.

\section{Models and Runtime}\label{app:models}

{\raggedright DeepSeek-V4.1-Flash was accessed through the DeepSeek API under the alias \texttt{deepseek-flash} in September and October 2026, and Claude Sonnet 4 through OpenRouter as \texttt{anthropic/claude-sonnet-4} in September 2026. Qwen3.8-27B ran locally in a 5-bit GGUF quantization with llama.cpp (build b11146), in non-thinking mode, with a 16{,}384-token context and a fixed seed. Llama-3.1-8B-Instruct, used only as the drafter in Section~\ref{sec:analysis}, ran on the same server with the same settings in a 5-bit quantization. All models used temperature 0 and at most 1{,}024 output tokens. Budget token counts use the same local tokenizer, DeepSeek's, for every model. Local runs used one NVIDIA RTX 5090 Laptop GPU with 24 GB of memory, and all API calls made for this study, including development runs, cost about USD 101.\par}

\end{document}